\documentclass[10pt,twocolumn,letterpaper]{article}

\usepackage[pagenumbers]{cvpr}

\usepackage[dvipsnames]{xcolor}
\definecolor{cvprblue}{rgb}{0.21,0.49,0.74}
\usepackage[pagebackref,break links,color links,citecolor=cvprblue]{hyperref}
\usepackage{bm}
\usepackage{color}
\usepackage{floatrow}
\usepackage{multirow}
\newfloatcommand{capbtabbox}{table}[][\FBwidth]

\title{
Cinematic  Behavior Transfer via NeRF-based Differentiable Filming
\vspace{-10pt}
}

\author{
	Xuekun Jiang$^{1*}$ \:
	Anyi Rao$^{2*}$ \:
	Jingbo Wang$^{1}$ \:
	Dahua Lin$^{1,3}$ \:
	Bo Dai$^{1}$ \\
	\small{
		$^1$Shanghai Artificial Intelligence Laboratory
		\quad
		$^2$Stanford University\quad
		$^3$The Chinese University of Hong Kong
	} \\
	\small{\texttt{\{jiangxuekun, wangjingbo, daibo\}@pjlab.org.cn}}
	\vspace{-3pt} \\
	\small{\texttt{anyirao@cs.stanford.edu}} \quad
	\small{\texttt{dhlin@ie.cuhk.edu.hk}} 
	\\
		$*$ denotes equal contribution
}

\begin{document}

\twocolumn[{
\renewcommand\twocolumn[1][]{#1}
\maketitle
\begin{center}
    \captionsetup{type=figure}
	\vspace{-44pt}
	\includegraphics[width=\textwidth]{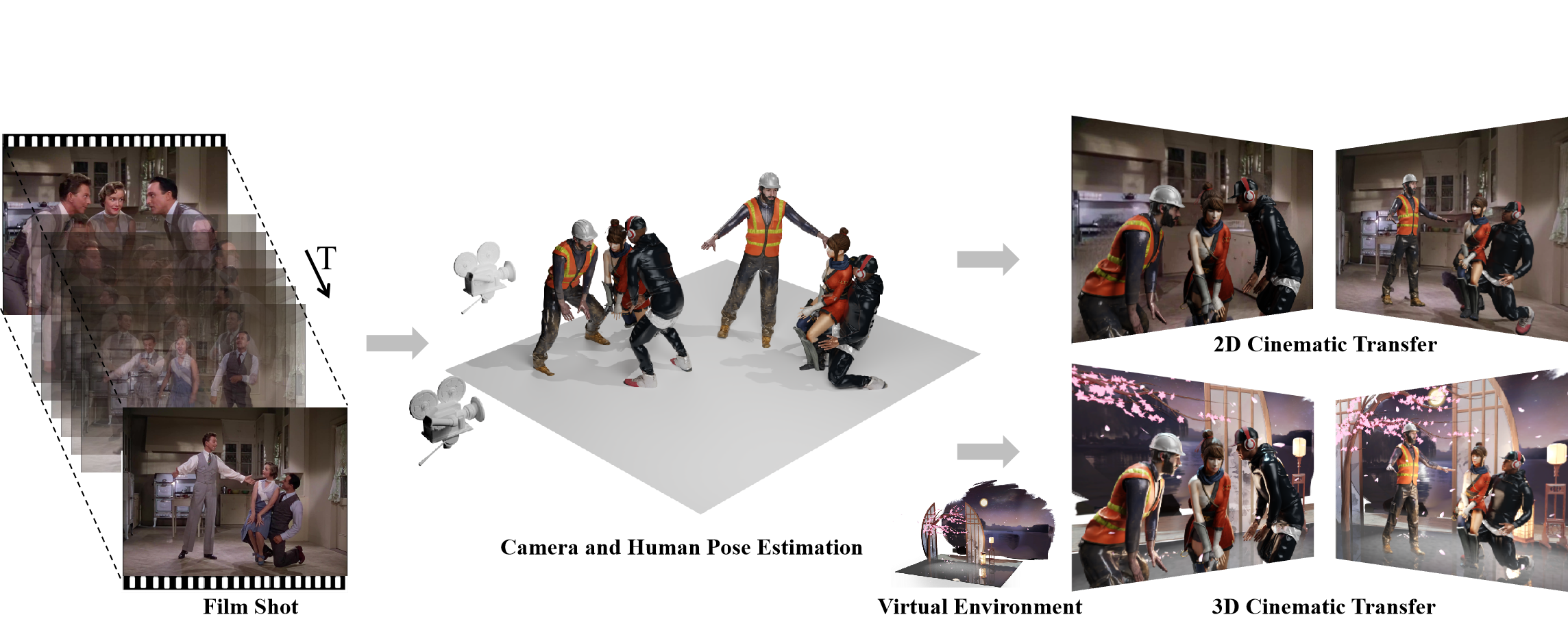}
	\vspace{-20pt}
	\captionof{figure}{ 
	Given a film shot, a series of visual continuous frames, containing complex camera movement and character motion, 
	we present an approach that estimates the camera trajectory and character motion in world coordinates. 
	The extracted camera and characters' behavior can be applied to new 2D/3D content through our cinematic transfer pipeline. 
	2D cinematic transfer aims to substitute the characters in the original shot with new 3D characters while preserving the identical character motion and camera movements. 
	3D cinematic transfer can apply the movements of characters and cameras to new characters and scenes, 
	providing more flexibility to modify various properties, such as lighting, character motion, and camera trajectory. 
	}
    \label{fig:teaser}
\end{center}
}]

\begin{abstract}
\vspace{-10pt}
In the evolving landscape of digital media and video production, the precise manipulation and reproduction of visual elements like camera movements and character actions are highly desired. 
Existing SLAM methods face limitations in dynamic scenes and 
human pose estimation often focuses on 2D projections, neglecting 3D statuses. 
To address these issues, we first introduce a reverse filming behavior estimation technique. 
It optimizes camera trajectories by leveraging NeRF as a differentiable renderer and refining SMPL tracks. 
We then introduce a cinematic transfer pipeline that is able to transfer various shot types to a new 2D video or a 3D virtual environment. 
The incorporation of 3D engine workflow enables superior rendering and control abilities, which also achieves a higher rating in the user study.
\vspace{-10pt}
\end{abstract}

\section{Introduction}
\label{sec:intro}
With the ongoing increase in media consumption, 
creators are consistently exploring innovative techniques to enhance the viewing experience, reduce production costs, and create compelling narratives.
Thus, the ability to manipulate and reproduce specific visual elements, such as camera movements and character behaviors, has long been a sought-after capability in the realm of digital media and video production,
as it helps maintain continuity and transfer a particular style or a unique mood from one scene to another.

It is challenging to replicate specific camera and character behaviors across different video clips.
Manually achieving it can be both time-consuming and prone to inconsistencies. 
Alternatively,
one can adopt SLAM~\cite{durrant2006simultaneous,schoenberger2016sfm,schoenberger2016mvs} and SMPL estimation~\cite{loper2023smpl} to respectively recover camera poses and human poses,
yet they struggle to handle complex scenarios with both camera and character behaviors.
Although SLAM and SMPL estimation can be used together~\cite{pavlakos2022one, ye2023decoupling, kocabas2024pace} to infer both camera and character behaviors,
it still has problems of mismatch with the original shot due to the noise caused by dynamic content.
%
Recently, some researchers~\cite{yen2021inerf,wang2023jaws} took advantage of NeRF~\cite{mildenhall2020nerf} to inverse-optimize camera poses with fewer effects from dynamic content.
Nevertheless, they required manual preparation of a similar scene as NeRF training data, which significantly diminishes their flexibility and scalability.

To address the above challenges, 
we propose a filming behavior transfer pipeline, 
that estimates SMPL tracks and camera trajectory of a film shot.
The SMPL track is a sequence of human pose.
%
We detect the character motion in reference shot and approximate it by predicting SMPL tracks.
While SMPL tracks can be estimated as in previous methods~\cite{ye2023decoupling}, 
we train a dynamic NeRF~\cite{pumarola2021d} to represent the 3D SMPL tracks. 
We then take NeRF as a differentiable renderer to provide image-level matching supervision for camera trajectory optimization.
%
The optimized camera trajectory can further refine the SMPL tracks, 
leading to a more accurate estimation of character and camera behaviors.
As the above SMPL tracks do not have textures, 
to meet the artist's workflow needs, 
we further develop a 3D engine-based workflow to adapt the SMPL track and camera to new virtual characters, 
enabling a higher level of control and precision in the creative process, such as changing the lighting or adjusting the speed of the camera movement.
With this, we can transfer a variety of shot types, 
including different shot scales, angles, complex camera movements, and various character numbers,
which helps artists create new content with similar cinematic behavior.

Extensive experiments show the capacity of our method to extract reasonable character motions and camera trajectory from a given well-known movie shot and generate new content with a similar cinematic style through a 3D engine workflow.

\section{Related Work}
\label{sec:related}

\noindent\textbf{Human and Camera Motion Estimation.}
Extracting human and camera motion from video has attracted increasing attention from researchers in recent years.
Most recent methods~\cite{kanazawa2019learning, choi2021beyond} were just focused on how to estimate the human pose in 3D because of the fixed camera.
For dynamic camera trajectory, some approaches~\cite{li2022d, yu2021human,yuan2022glamr} have tried to circumvent the issue of camera motion by recovering the human trajectories in global coordinates from the per-frame local human poses.
Other researchers~\cite{liu20214d, henning2022bodyslam, pavlakos2022one, ye2023decoupling, kocabas2024pace} have introduced SLAM system into human pose estimation to reconstruct the 4D human pose.
Pavlakos~\etal~\cite{pavlakos2022one} proposed a method to reconstructed 3D humans and environments in TV shows.
They used COLMAP and NeRF to reconstruct the cameras and dense scene and use this information to recover accurate 3D pose and location of people over shot boundaries and on monocular frames.
Ye~\etal~\cite{ye2023decoupling} proposed a method to reconstruct global human trajectories from videos in the wild.
They showed that relative camera estimates along with data-driven human motion priors can resolve the scene scale ambiguity and recover global human trajectories.
Kocabas~\etal~\cite{kocabas2024pace} proposed to tightly integrate SLAM and human motion priors in an optimization that is inspired by bundle adjustment.
Unlike the above methods that used SLAM as initialization, our method optimizes camera trajectories by leveraging NeRF as a differentiable renderer.

\vspace{2pt}
\noindent\textbf{NeRF-based Camera Pose Estimation.}
NeRF~\cite{mildenhall2021nerf} is a popular representation of 3D scenes, which uses a multilayer perceptron (MLP) that evaluates a 5D implicit function estimating the density and radiance emanating from any position in any direction.
Yen~\etal~\cite{yen2021inerf} first proposed to estimate mesh-free camera pose by ``inverting" a NeRF.
A lot of recent work~\cite{meng2021gnerf,lin2021barf,chng2022garf,levy2023melon,smith2023flowcam,cheng2023lu} focused on how to get camera parameters without using SFM, and instead train both camera parameters and NeRF during training using only pictures.
Most of them cared more about the quality of the NeRF than the quality of the camera.
And iNeRF~\cite{yen2021inerf} is limited to its slow inference speed and it's very sensitive to the initial parameters of the camera.
To address these problems, 
Lin~\etal~\cite{lin2023parallel} improved it via 1) using Instant NGP to replace the native NeRF~\cite{muller2022instant}; 
2) introducing parallel Monte Carlo sampling to overcome local minima and improved efficiency in a more extensive search space of camera parameters.
Wang~\etal~\cite{yoo2021virtual}  proposed a feature-driven cinematic motion transfer technique. 
It replicated the camera sequences from movies to a trained NeRF to let the generated video clip maximize the similarity with the reference clip through a designed cinematic loss.
Most recent works required manual preparation of a similar scene as NeRF training data.
Our approach predict the SMPL tracks from original shot and use a dynamic NeRF to represent it.
Hence, our approach eliminates the need for manual scene construction.

%
%
%
%
%
%
%

\section{Method}
\label{sec:method}
This paper proposes a method to transfer character and camera behavior from a given single shot to new 2D/3D content.
For each shot, we extract its SMPL tracks representing the sequence of characters' motion in world coordinates, 
and then optimize the camera trajectory based on the SMPL tracks, as detailed in Sec.~\ref{sec:pipeline} and Sec.~\ref{sec:camera}.
Finally, we used the SMPL tracks and the camera trajectory to create new content through a full 3D engine workflow in Sec.~\ref{sec:transfer}.

\begin{figure*}[tp]
    \centering
    \includegraphics[width=\linewidth]{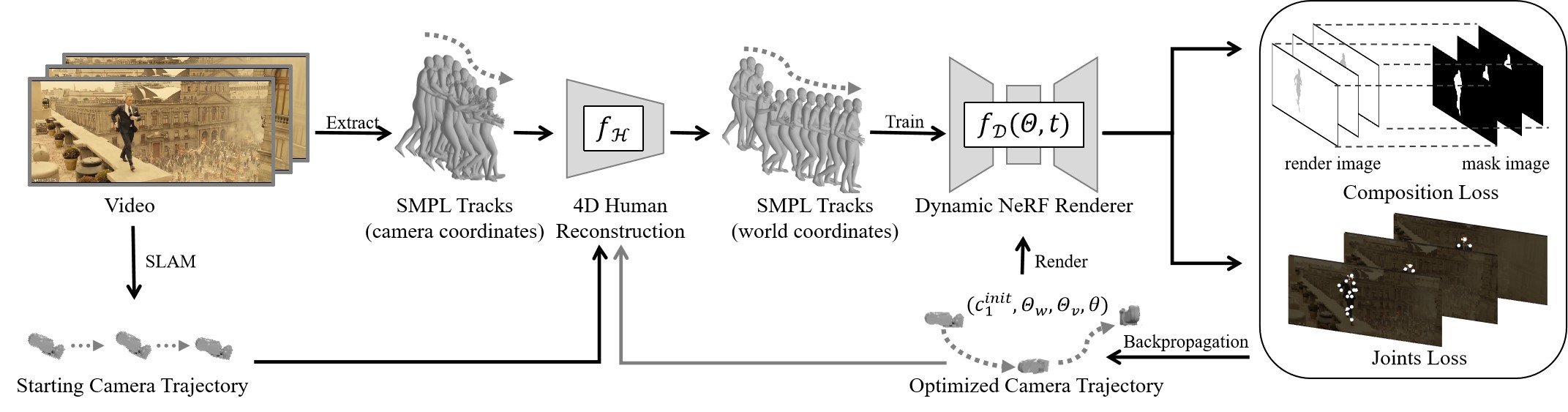}
    \vspace{-13pt}
    \caption{The overview of our differentiable pipeline of characters' motion and camera trajectory estimation. 
    Given a shot video, we first extract SMPL tracks in camera coordinates and a camera trajectory in world coordinates. 
    Then, we reconstruct the motions of all characters in world coordinates by a 4D human reconstruction method. 
    Finally, we optimize camera trajectories by leveraging NeRF as a differentiable renderer and refine SMPL tracks.}
    \label{fig:pipeline}
    \vspace{-11pt}
\end{figure*}

\subsection{Human Pose and Camera Estimation}
\label{sec:pipeline}
Due to the coupling of character and camera movement in a video, it is hard to obtain accurate human motions in world coordinates. 
We need to estimate a camera trajectory in the world coordinates to decouple the human and camera motions to eliminate the ambiguity of scene scale in camera space. 
Typically, we utilize SLAM (simultaneous localization and mapping) to extract the camera trajectory.
Given a single shot $V$ with $T$ frames, $V = \{I_1, \dots, I_T\}$ with $N$ characters, 
we first predict a starting camera trajectory $\hat{C} = \{ \hat{c}_t \}_{t=1}^T$ in world coordinates with SLAM method.
And then we 
predict $N$ SMPL tracks $S_{c} = \{ S_{c,n}\}_{n=1}^N$ in camera coordinates.
%
%
With $S_{c}$ and $\hat{C}$,
we can compute the SMPL tracks $S_{w}$ in world coordinates with a 4D human reconstruction method $f_{\mathcal{H}}$,
\begin{equation}
S_{w}=f_{\mathcal{H}}(S_{c}, \hat{C}).
\label{eq:smpltracks}
\end{equation}
Although the starting camera trajectory $\hat{C}$ predicted from $V$ can resolve the scene scale ambiguity and help us to recover the human motions in world coordinates,
$\hat{C}$ suffers from the errors caused by the dynamic content in $V$.
We propose to optimize a new camera trajectory $C^\ast=\{ c^\ast_t \}_{t=1}^T$ based on $S_{w}$ via a differential render NeRF to add image-level supervision.

NeRF represents a 3D scene in a differentiable way that can render an image with a given camera pose. 
This also means that, for a trained NeRF, we can figure out an optimal camera pose that renders frames that match our reference shot $V$.
Towards this goal, we train a dynamic NeRF $f_\mathcal{D}(\Theta, t)$ to capture the character motion tracks $S_{w}$.
Hence, the camera trajectory optimization can be treated as an inverted optimization, 
which takes a trained NeRF as a differentiable render and uses the NeRF backpass gradients to find the optimal camera parameters.
\begin{equation}
c^{\ast}_t=\underset{c_t\in\text{SE}(3)}{\mathrm{argmin}} \mathcal{L}(c_t \ | \  I_t,\Theta),
\end{equation}
where $\Theta$ is the parameters of NeRF,  $I_t$ is the reference shot frame image in time $t$.
Due to the lack of background and details of the approximation scene of SMPL, our method can not compute the loss directly from the RGB images like iNeRF~\cite{yen2021inerf}, and optical flow or human pose like JAWS~\cite{wang2023jaws}.
To tackle the above issues, 
we introduce two losses: a composition loss $\mathcal{L}_c$ and a joint loss $\mathcal{L}_j$. 
i) The composition loss $\mathcal{L}_c$ is calculated by an instant mask image for the original image and the NeRF rendered image.
For the mask images, We color the pixels of each human mask to match the vertex color of the corresponding SMPL model and color the rest of pixels white.
ii) The joint loss $\mathcal{L}_j$ is calculated by the joint distances between the original image and the rendered image. 
For the original image, we predict the joints 2D coordinates of each character as the ground truth by ViTPose~\cite{xu2022vitpose}.
For the rendered image, we reproject the SMPL joints in 3D to 2D by optimized camera pose.
The final loss is a weighted sum of $\mathcal{L}_c$ and $\mathcal{L}_j$.

After the optimization, we obtain a more accurate camera trajectory $C^\ast$. 
Compared to the starting camera trajectory $\hat{C}$ predicted solely by SMPL, 
it can better align the reconstructed image framing with the original shot 
by using NeRF as the differentiable renderer with two image-level losses.
Since a more reasonable camera trajectory leads to better SMPL tracks in world coordinates,
we can update Eq.~\eqref{eq:smpltracks} and compute more accurate SMPL tracks,
\begin{equation}
S^\ast_{w}=f_{\mathcal{H}}(S_{c}, C^\ast).
\end{equation}

\subsection{Camera Trajectory Optimization}
\label{sec:camera}
A key step in the above optimization is to optimize the camera trajectory.
Nevertheless, using NeRF to directly regress the values in the transformation matrix $c^\ast$ 
does not guarantee that the optimized result remains in the $SE(3)$ manifold.

\vspace{1pt}
\noindent\textbf{Preliminary on optimiazation parameters.}
In most NeRF-based pose estimation works~\cite{yen2021inerf}, 
the camera pose is defined as a transform matrix $c^\ast \in SE(3)$ in world coordinates, 
and is estimated by a trained NeRF model.
To ensure the estimated pose still lies in the $SE(3)$ manifold during gradient-based optimization, 
the camera pose $c^\ast$ is represented by an initial pose estimate $c^{\text{init}} \in SE(3)$ and a transformation matrix $A$ with exponential coordinates:

\begin{equation}
    \label{eq:single_cam}
   \begin{aligned}
    &c^\ast = A c^{\text{init}}, \\
    &\text{where} \ A=e^{[S]\theta}=\begin{bmatrix}
     e^{[w]\theta  } & f_{\mathcal{K}(\theta ,w,v)} \\
      0&1
    \end{bmatrix}, \\
    &f_{\mathcal{K}(\theta ,w,v)}=(I\theta+(1-\cos{\theta})[w]+(\theta-\sin{\theta}[w]^2))v.
    \end{aligned} 
\end{equation}
Here $S=[w,v]^T$ represents the screw axis, $\theta$ is a magnitude, $[w]$ represents the skew-symmetric 3 × 3 matrix of $w$. 
The matrix optimization problem is then equivalent to figuring out the optimal parameters $(\theta, w, v)$.
With this parameterization according to Eq.~\eqref{eq:single_cam}, the optimal goal is to achieve optimal relative transformation from an initial estimated pose $c^{\text{init}}$:
\begin{equation}
\label{eq:single-camera-param}
    \theta, w, v = 
    \underset{S\theta\in\mathbb{R}^6 }{\mathrm{argmin}}\mathcal 
    \ 
    \mathcal{L}(Ac^{\text{init}} \ | \ I, \Theta ).
\end{equation}
For each given observed image, the camera parameters $(\theta ,w,v)$ are initialized near $0$, and each is drawn from random from a zero-mean normal distribution $\mathcal{N}(0,\sigma = 10^{-6})$.

%
%
%

\vspace{2pt}
\noindent\textbf{Sequence camera parameters optimization.}
Existing NeRF-based camera pose estimation works~\cite{yen2021inerf, wang2023jaws} focus on single-camera pose estimation and rarely tackle a sequence.
Their primary objective is to ensure that the rendered image resulting from camera optimization closely resembles the target image. Consequently, they often prioritize this visual similarity over the accuracy of the trajectory in 3D world coordinates.
They optimized each time camera parameter $c^\ast_t$ independently during this process.

We aim to optimize the camera pose to produce a correct 2D image and predict the correct inverted 3D camera trajectory.
Regrettably, constructing a 3D scene and obtaining a NeRF representation identical to that of a movie shot with complex dynamic content significant challenges.
We accurately detect the pose of the character in the reference shot and subsequently estimate it by predicting SMPL tracks.
However, the introduction of noise through SMPL predictions amplifies errors in camera pose estimation, leading to an unreasonable camera trajectory in 3D world coordinates.
These motivate us to learn a continuous representation of the camera trajectory to prevent mutations.

As mentioned in Eq.~\eqref{eq:single_cam}, 
the camera trajectory parameters $c^\ast_t \in C^\ast$  for each time step $t$
can be decomposed by an initial pose $c^{\text{init}}_t$ and a transformation matrix $A_t$
 \begin{equation}
 \label{eq:baseT}
     c^\ast_t  = A_tc^{\text{init}}_t.
 \end{equation}
To learn a continuous representation of the camera trajectory, 
we use two strategies.
First, the initial pose $c^{\text{init}}_t$ is derived from the camera parameters of the preceding moment: $c^{\text{init}}_t = c^\ast_{t-1}$. 
Second, we defined a continuous function $f_\mathcal{A}$ with respect to time $t$ to calculate transformation matrix $A_t$: $A_t = f_\mathcal{A}(t)$.
The matrix $A_t$ is defined by parameters $(\theta ,w_t,v_t)$ according to Eq.~\eqref{eq:single_cam}.
The parameter $\theta$  remains constant throughout the entire camera trajectory Therefore, as long as the parameters $w_t$ and $v_t$ are continuous, the continuity of the camera trajectory can be guaranteed.
We use two MLP $f_\mathcal{W}$ and $f_\mathcal{V}$ to predicted the $w_t$ and $v_t$ in each time step $t$:
\begin{equation}
\begin{aligned}
    w_t & = w_1 + f_\mathcal{W}(t), \\
    v_t & = v_1 + f_\mathcal{V}(t).
\end{aligned}
\end{equation}
As mentioned in Sec.~\ref{sec:pipeline}, the $S_w$ is projected to the world coordinates by SLAM camera $\hat{C}$, and then we used the $S_w$ to optimize the camera trajectory $C^\ast$.
The camera trajectory $C^\ast$ is aligned to the SLAM camera $\hat{C}$.
To calculate the parameters $\theta$, $w_1$ and $v_1$ in the first time step.
So we can use the $\hat{c}_1$ as the $c^{\text{init}}_1$ to optimise the first camera pose $c^{\ast}_1$ according to Eq.~\eqref{eq:single_cam} that can reduce the computational cost. 
We use Eq.~\eqref{eq:single-camera-param} to optimize the $\theta$, $w_1$ and $v_1$.

Finally, we transform Eq.~\eqref{eq:baseT} into the following:
 \begin{equation}
    c^\ast_t  = f_\mathcal{A}(\Theta_w, \Theta_v, \theta, t)c^\ast_{t-1}, \quad t \in [2, t),
\label{eq:cameratraj}
 \end{equation}
where $\Theta_w$ and $\Theta_v$ is the parameters of MLP $f_\mathcal{W}$ and $f_\mathcal{V}$.

\begin{figure*}[tp]
    \centering
    \includegraphics[width=\linewidth]{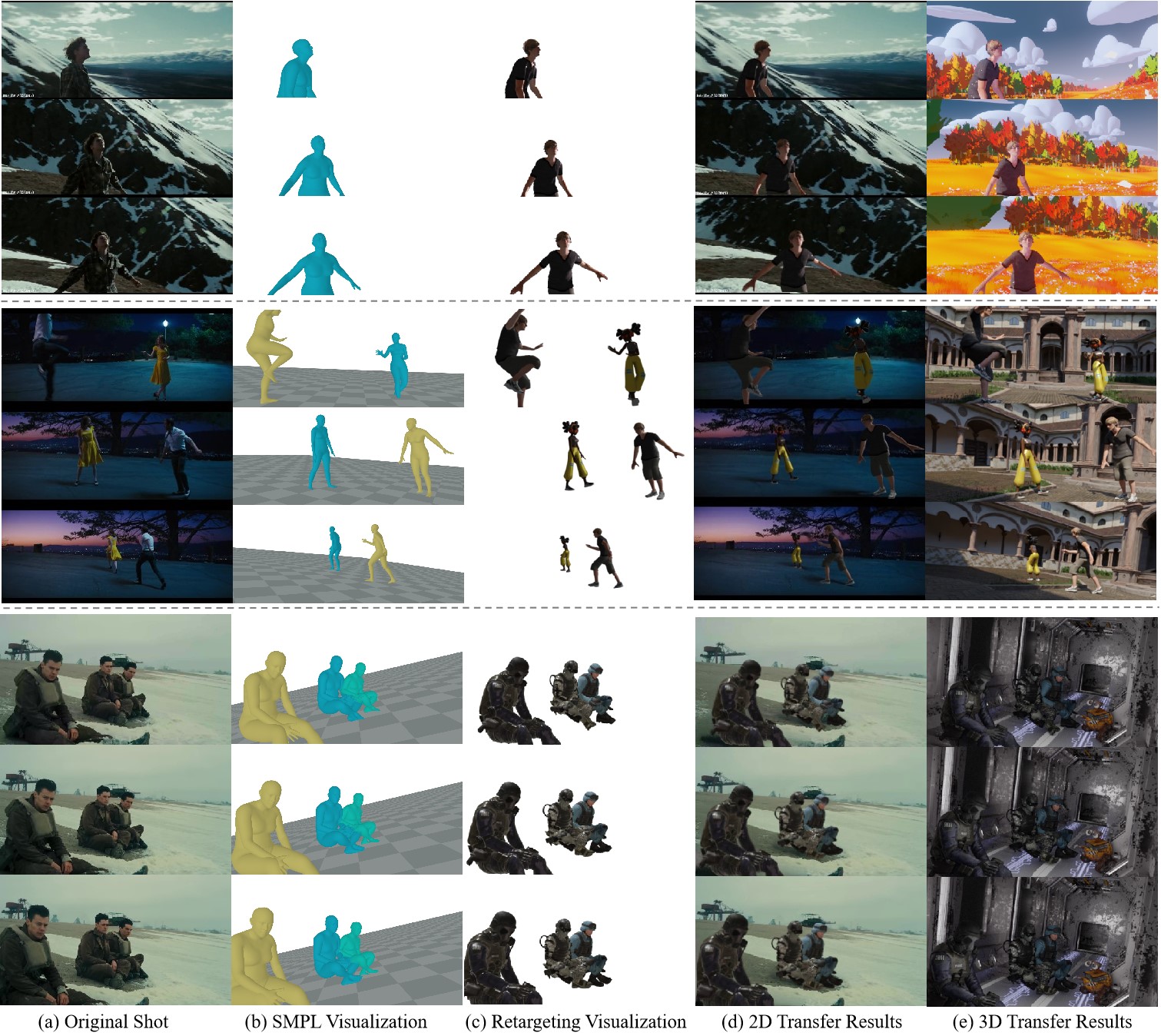}
    \vspace{-10pt}
    \caption{Examples of cinematic behavior transfer. (a) The original shot: we present three common shot types Arc, Track, and Push In. (b)  SMPL visualization of our method. We recreate the cinematic behavior by extracting the SMPL tracks and the camera trajectory from the original shot. (c) New Characters retargeting visualization of our method. We apply the motion of SMPL to new character and render images with optimized camera trajectory through our engine workflow. 
    (d) The 2D cinematic transfer results. 
    We erase the characters in the foreground and combine the background video with (c) to generate a new 2D video like (a). 
    (e) The 3D cinematic transfer results. 
    We apply the motion and camera to a new virtual scene, like a cartoon grassland, palace, and SF tunnel.}
    \label{fig:results}
    \vspace{-10pt}
\end{figure*}

\subsection{Transfer via a 3D Engine-based Workflow}
\label{sec:transfer}
With the accurate character and camera behavior estimation, we can transfer them to 2D and 3D content which are shown in Fig.~\ref{fig:results} (d) and (e).

2D cinematic transfer aims to replace characters in an existing film shot with new 3D characters, while keeping the same character motions and camera behavior. 
For 2D cinematic transfer, we first render a pure video without background $V_{f}$ with our camera trajectory and the character after retargeting.
Then, we remove the foreground characters from the original shot. 
In this paper, we use an advanced object removal method ProPainter~\cite{zhou2023propainter}, to erase the characters and generate a pure background video $V_{b}$.
We combine these two videos $V_{f}$ and $V_{b}$ to obtain the final results.

3D cinematic transfer takes character and camera movements and applies them in new characters and scenes,
which further allows for adjustments in different aspects like lighting, character motion, camera motion, offering more control and options for personalizing the final result.
3D cinematic transfer is much simpler, which only needs to apply the motion and camera to the virtual scene and render the result.
However, compared to the 2D workflow, 
3D cinematic transfer has more flexibility.
Since the entire scene is in the 3D space, we are free to modify it according to our needs, for example, change the time from night to day (Fig.~\ref{fig:results} middle (e)), place a robot in the corner (Fig.~\ref{fig:results} below (e)).

%
%
%

%

\section{Experiments}
\label{sec:exp}
\subsection{Implementation Details.}
Our implementation is based on `torch-ngp' and Pytorch.
We use PHALP~\cite{rajasegaran2022tracking} for human pose tracking, SLAHMR~\cite{ye2023decoupling} for 4D human reconstruction, D-NeRF~\cite{pumarola2021d} for neural rendering and VitPose~\cite{xu2022vitpose} for 2D joints prediction.
More details are shown in the supplementary.
%
%

\begin{figure}[tp]
    \centering
    \includegraphics[width=\linewidth]{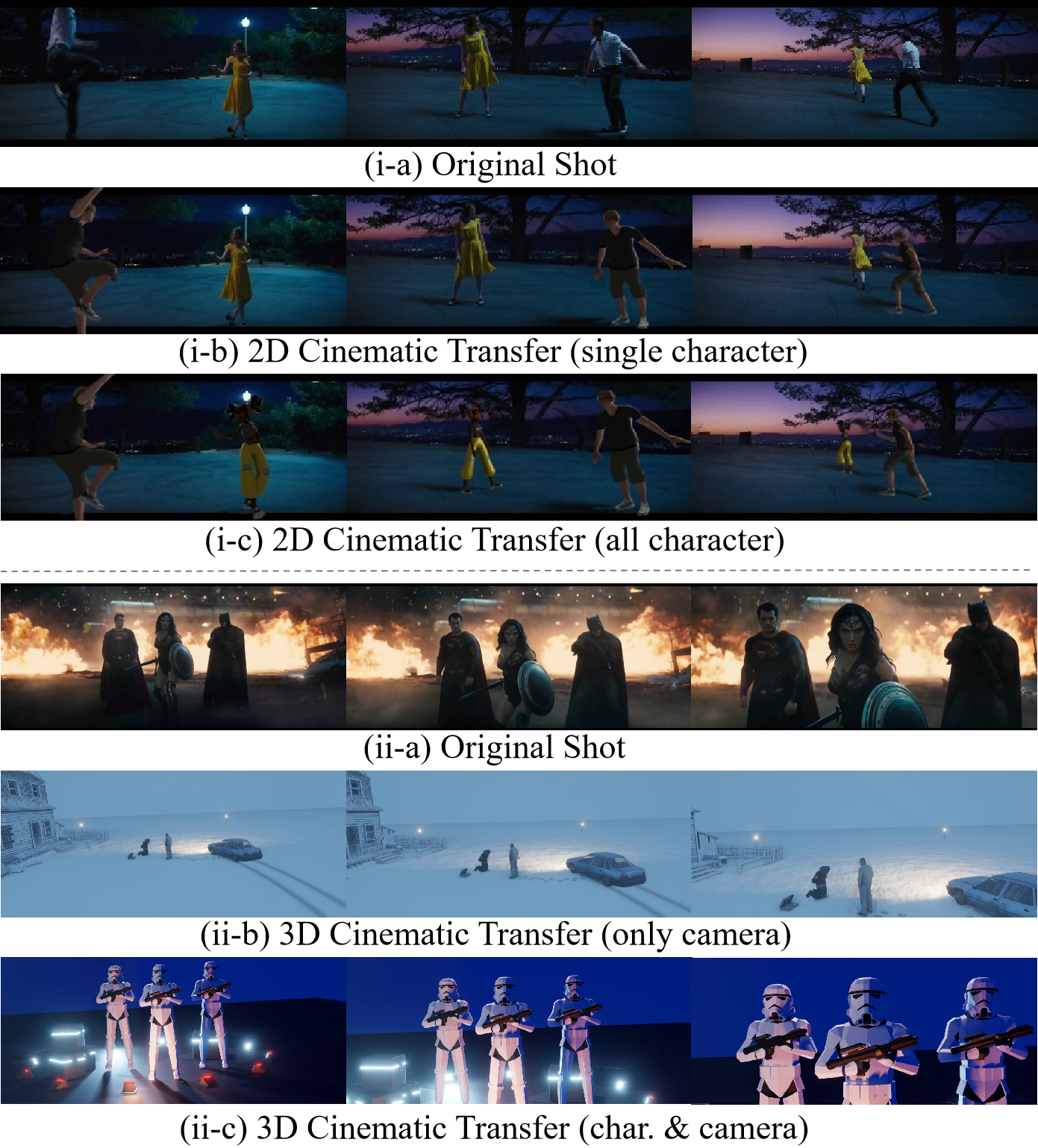}
    \vspace{-10pt}
    \caption{
    Flexibility enabled by 3D engine workflows.
    For 2D cinematic transfer, we can replace any character like (i-b) or (i-c), in shot (i-a).
    For 3D cinematic transfer, we demonstrate the ability to apply either the camera alone (ii-b) or both character and camera information to new content(ii-c).}
    \label{fig:transfer}
    \vspace{-10pt}
\end{figure}

\subsection{Qualitative Results}
Our approach can reproduce character movements and camera trajectories from a given shot, replace characters in the frame, or change the scene through an engine-based workflow. 
Fig.~\ref{fig:results} showcases some examples of cinematic transfer from an original film shot to a synthetic environment.
Compared with other methods, 
our method can handle various scenes with single/double/multiple characters.
It can also recover various types~\cite{rao2020unified, rao2023dynamic} of shot trajectories, such as Arc, Track, and Push-In.

Specifically, in Fig.~\ref{fig:results}: 
(a) The original shot: we present three common shot types Arc, Track, and Push-In.
Arc shot orbits the camera around a subject in an arc pattern.
Track shot moves the camera through the scene for an extended amount of time.
Push-In shot moves the camera closer to a subject.
(b) SMPL visualization of our estimation. 
We extract and optimize the SMPL tracks and the camera trajectory from the original shot, 
which can reproduce the characters' motion and the camera movement.
(c) New 3D characters' retargeting visualization.
We retarget the original shot's SMPL motion  to new 3D characters and 
render images with optimized camera trajectory through our engine workflow.  
This allows us to further put them into a scene.
(d) The 2d cinematic transfer results. 
We replace any character in the original shot.
For instance, 
we can replace ``Mia" and ``Sebastian", the characters in \emph{Lalaland}, with our own 3D characters or keep ``Mia" and replace ``Sebastian" with a new 3D character as shown in Fig.~\ref{fig:transfer} (i-b).
It is implemented by first using an advanced object removal method~\cite{zhou2023propainter} to erase the character in the foreground and combining the background video with (c) to generate the final video. 
(e) The 3d cinematic transfer results. 
We apply the characters' motion and camera movement to new 3D characters and a new virtual scene. 
This provides the user with more freedom of operation, 
\emph{e.g.}, adjusting the lighting, or modifying the camera.

\vspace{1pt}
\noindent\textbf{Flexibility of our cinematic transfer.}
Fig.~\ref{fig:transfer} shows various transfer results with our 3D engine-based workflow. 
After investigating artist's workflows, 
we figured out that to truly meet their needs, 
the workflow should enable freely using different information extracted from the video, 
such as the motions of the characters or the movement of the camera.
Our approach can well provide this flexibility to explore various creative possibilities: replacing one character and keeping the others within a movie shot (Fig.~\ref{fig:transfer} (i-c)); 
employing solely the camera trajectory from a movie shot to a new scene with different character motions (Fig.~\ref{fig:transfer} (ii-b)).
However, the recent SOTA cinematic transfer method JAWS~\cite{wang2023jaws} does not fully support this flexibility. 
Their workflow can only apply camera trajectory to scenes that closely resemble the reference video in terms of the number of characters and their relative positions.

\begin{figure}[tp]
    \centering
    \includegraphics[width=\linewidth]{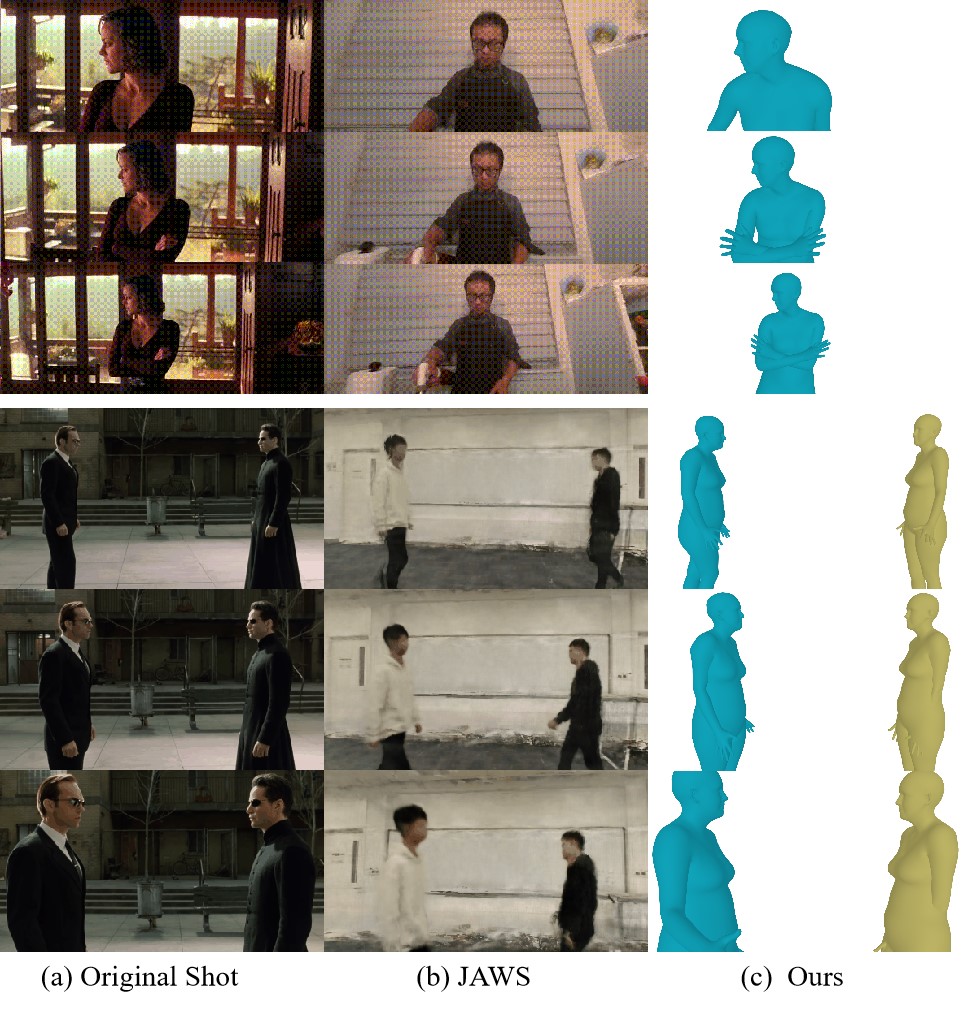}
    \vspace{-10pt}
    \caption{Comparison of the cinematic transfer results with SOTA. Our method demonstrates a better ability to align with the composition of the original shot compared to JAWS~\cite{wang2023jaws}.
    The two shots are from \emph{Inception, 2010} (top) and the \emph{Matrix, 1999} (below)
    }
    \label{fig:results_jaws}
    \vspace{-15pt}
\end{figure}

\vspace{1pt}
\noindent\textbf{Comparison with SOTA in cinematic transfer.}
JAWS~\cite{wang2023jaws} is an optimization-driven approach that addresses the cinematic transfer from a reference clip to a trained NeRF.
It used a \emph{on-screen} loss and a \emph{iner-frame} loss to cover the framing and camera motion aspects.
It is limited to handling highly mismatched character poses due to inter-frame motion.
So, it has to hand-craft a scene that is essentially the same as the original shot to train NeRF, which greatly limits its usage scenarios.
Additionally, simply adapting JAWS to our setting does not work due to the lack of background and details of the predicting SMPL tracks.
To be specific, the RAFT~\cite{teed2020raft} they used for the optical flow estimation cannot work for the rendered SMPL image
and LitePose~\cite{wang2022lite} they used to infer the post joint will ignore the inter-frame motion.
To achieve a strong comparison with JAWS, we use the shots used in their papers' experiments, as shown in Fig.~\ref{fig:results_jaws}.
Our method restores not only the composition of the shot but also the action of the characters.
Although JAWS used a realistic environment similar to the original shot, 
it does not accurately reproduce the composition of the original shot.
We can clearly observe the flaws in the ``Matrix" example, where two characters occupy the left and right parts of the image when the camera is pushed to the end.
The composition of the characters in JAWS's results do not closely align with the original footage.
Another limitation of JAWS is that it highly relies on dynamic NeRF results, which will easily fail on the complex motion shots that dynamic NeRF cannot handle.
As our method does not require background details of the predicting SMPL tracks, our approach has lower quality requirements for NeRF's quality,
which leads to stronger robustness and better performance.

\begin{figure}[tp]
    \centering
    \includegraphics[width=\linewidth]{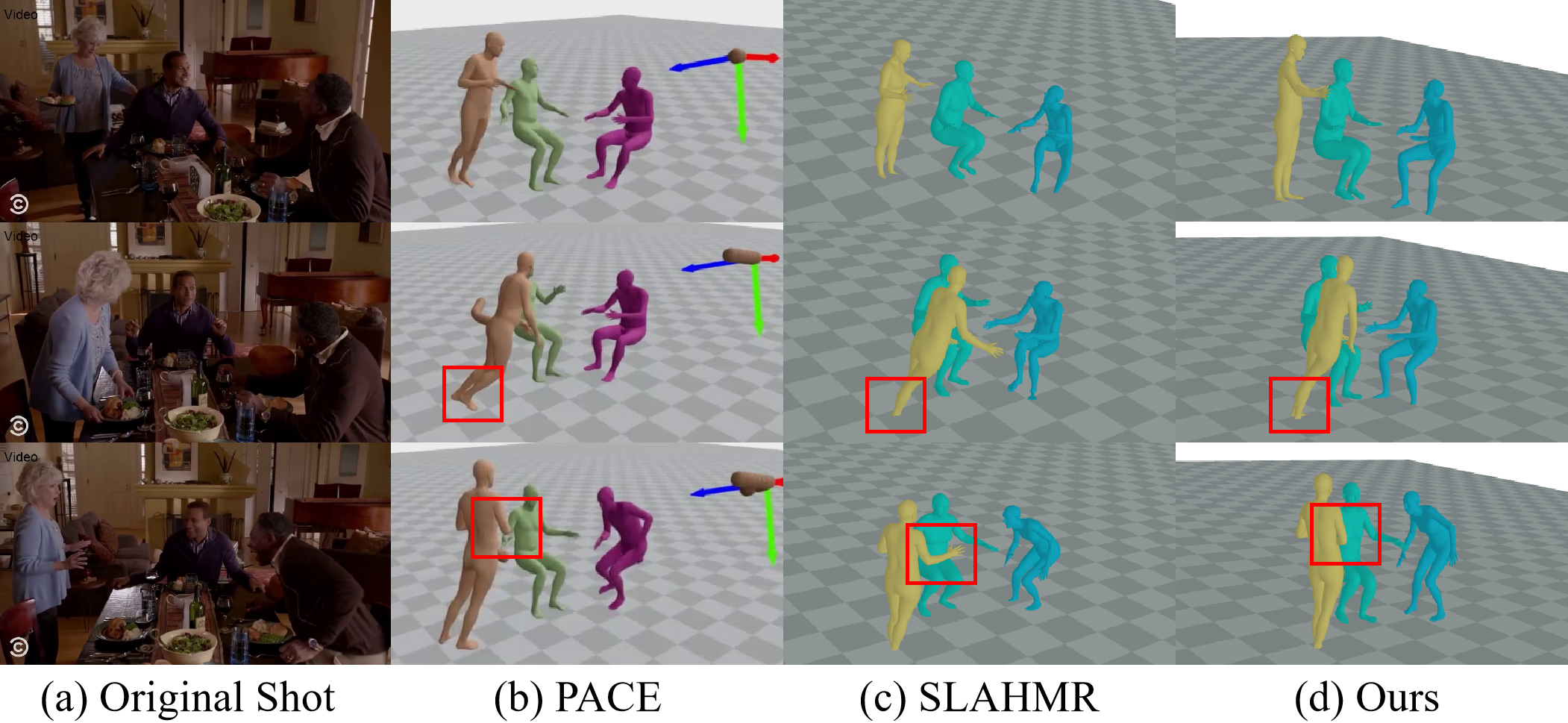}
    \vspace{-10pt}
    \caption{Comparison of the human pose estimation in world coordinates with SOTA. As shown in the red box, our method has better results in limb details due to the improved optimization of the camera trajectory.}
    \label{fig:results_smpl}
    \vspace{-10pt}
\end{figure}

\vspace{1pt}
\noindent\textbf{Comparison with SOTA in world coordinates human pose estimation.}
As we mentioned in Sec.~\ref{sec:pipeline}, when we finished the the camera $C^\ast$ optimization with $f_\mathcal{D}$, 
we can bring it into $f_\mathcal{H}$ and get new SMPL tracks.
Fig.~\ref{fig:results_smpl} compares the human pose estimation results with SOTA methods. 
Due to the enhanced optimization of the camera trajectory, 
our method achieves better results in capturing detailed limb poses.
Inspired by bundle adjustment, PACE~\cite{kocabas2024pace} tightly integrates SLAM and human motion priors in optimization. 
It can handle the shot with that entire character's body
but is limited to dealing with the shot where only the character's partial body appears.
To achieve a strong comparison with PACE, we use the shots used in their papers' experiments.
In Fig.~\ref{fig:results_smpl} (b), the feet of the figure on the left are not on the ground. 
Since films contain lots of partial body shots like close-up shot or medium shot, 
PACE is not very suitable for cinematic transfer.
SLAHMR~\cite{ye2023decoupling} used relative camera estimates along with data-driven human motion priors to resolve the scene scale ambiguity and recover the human trajectories in world coordinates.
However, the human pose is likely to fail due to the noise from DROID-SLAM on dynamic content.
In the leftmost character shown in the red box of Fig.~\ref{fig:results_smpl}~(c), 
the arms should be close to the body, while SLAHMR predicts open arms.
Although our method is based on SLAHMR, 
we do not directly use SLAM's camera trajectory and utilize NeRF as a differential render to re-optimize the camera pose.
Due to the improved optimization of the camera trajectory, our method achieves better results in capturing finer details of limb poses.

\begin{table*}[t!]
\centering
\resizebox{0.82\columnwidth}{!}{%
\begin{tabular}{c|ccc|ccc|ccc}
\hline
Shot Move. & \multicolumn{3}{c|}{PUSH-IN} & \multicolumn{3}{c|}{PULL-OUT} & \multicolumn{3}{c}{PAN} \\ \hline
Methods    & PA$\uparrow$      & IoU$\uparrow$      & MPJPE$\downarrow$  & PA$\uparrow$       & IoU$\uparrow$      & MPJPE$\downarrow$    & PA$\uparrow$     & IoU$\uparrow$    & MPJPE$\downarrow$   \\ \hline
DROID-SLAM~\cite{teed2021droid} & 86.2   & 85.8   & 404.9      & 86.0   & 85.3   & 356.2      & 91.9 & 89.6 & 40.9     \\
iNeRF~\cite{yen2021inerf}      & 89.0   & 88.2   & 292.6      & 92.8   & 91.5   & 83.9       & 83.9 & 81.0 & 109.6     \\
Ours       & \textbf{89.9}   & \textbf{88.5}   & \textbf{59.6}      & \textbf{94.8}   & \textbf{94.0}   & \textbf{23.8}       & \textbf{93.4} & \textbf{91.4} & \textbf{21.4}      \\ \hline
Shot Move. & \multicolumn{3}{c|}{TRACK}   & \multicolumn{3}{c|}{FOLLOW}   & \multicolumn{3}{c}{ARC} \\ \hline
Methods    & PA$\uparrow$      & IoU$\uparrow$     & MPJPE$\downarrow$  & PA$\uparrow$      & IoU $\uparrow$     & MPJPE$\downarrow$  & PA$\uparrow$     & IoU$\uparrow$    & MPJP$\downarrow$ \\ \hline
DROID-SLAM~\cite{teed2021droid} & 89.3   & 88.3   & 109.2      & 73.3   & 70.5   & 1046.9       & 92.7 & 92.6 & 145.2     \\
iNeRF~\cite{yen2021inerf}      & 90.1   & 89.2   & 58.5      & 85.3   & 85.1   & 267.5       & 90.8 & 90.5 & 116.3     \\
Ours       & \textbf{94.5}   & \textbf{93.8}   & \textbf{21.8}      & \textbf{91.3}   & \textbf{90.5}   &  \textbf{130.9}      & \textbf{94.8} & \textbf{94.5} & \textbf{47.9}  \\  \hline
\end{tabular}
}
\vspace{-8pt}
\caption{
{Comparison with the state-of-the-art camera pose estimation methods on different shot movement types.} 
Our approach outperforms the other baselines across all metrics.
}
\vspace{2pt}
\label{tab:constract}
\end{table*}

\begin{figure}[tp]
    \centering
    \includegraphics[width=\linewidth]{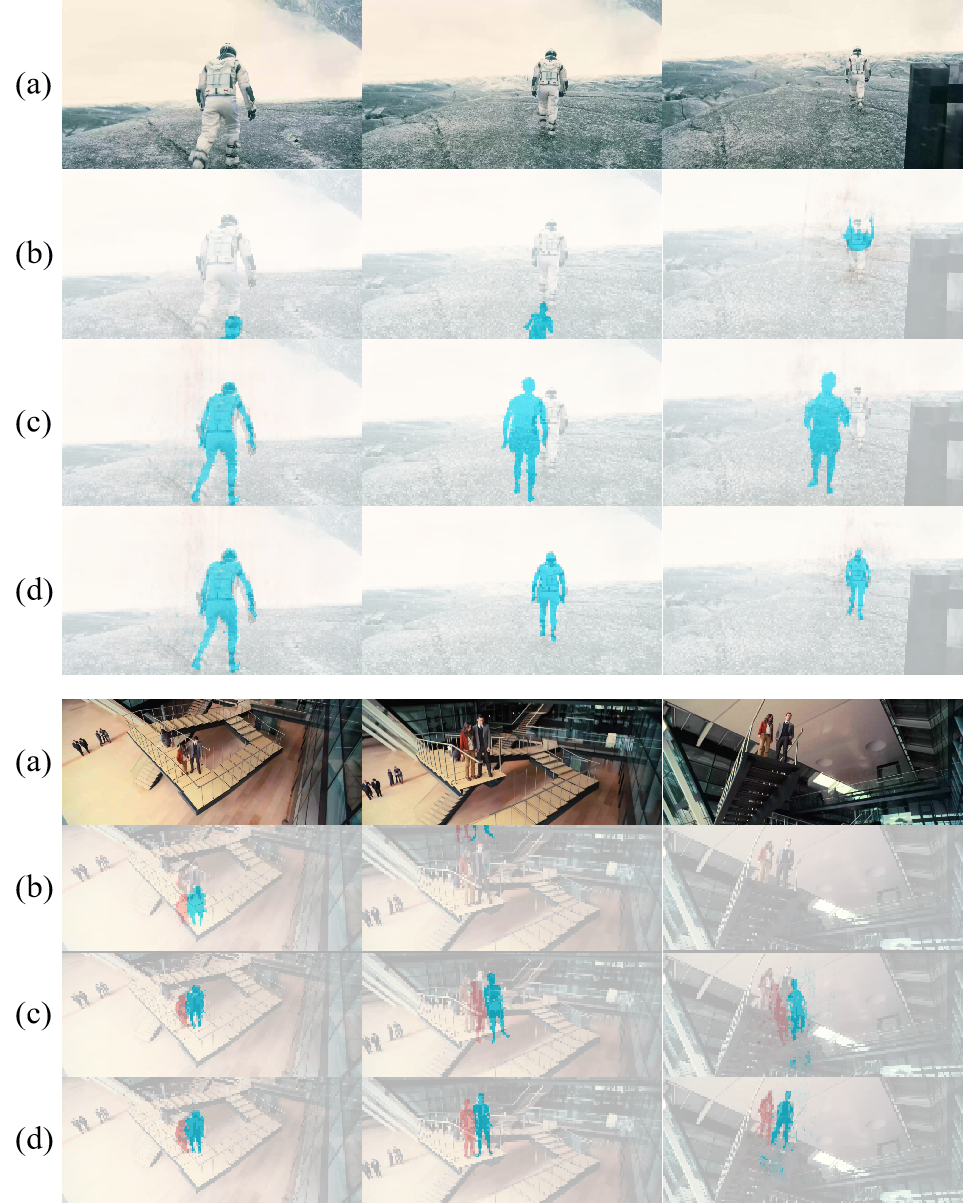}
    \vspace{-15pt}
    \caption{We show the results of the alignment visualization: (a) the original shot, (b) DROID-SLAM, (c) iNeRF, (d) our method.
    }
    \label{fig:results_compare}
    \vspace{-1pt}
\end{figure}

\subsection{Quantitative Results}
Since A key feature in cinematic behavior transfer is to check if the transferred results are similar to the original shot,
we test three metrics on the frame composition restoration and conduct a user study to validate users' satisfaction with 2D and 3D transfer.

\vspace{1pt}
\noindent\textbf{Comparison with SOTA on frame composition restoration.}
Tab.~\ref{tab:constract} shows the quantitative results between our method with two SOTA camera pose estimation methods DROID-SLAM~\cite{teed2021droid} and iNeRF~\cite{yen2021inerf}.
DROID-SLAM is a deep learning-based SLAM system with fewer catastrophic failures.
iNeRF uses NeRF for mesh-free, RGB-only 6DoF camera pose estimation, which uses RGB pixels as a supervision signal.
Due to the lack of background and details of the approximation scene of SMPL, we use the composition loss $\mathcal{L}_c$ instead of RGB loss to implement iNeRF.
%
We test more than 100 well-known film shots that collected from the Internet with multiple styles to show the effectiveness of our method.

To measure the effect of different methods on restoring picture composition, 
we use three metrics to evaluate all the methods: 
1) Pixel Accuracy (PA): It is the percentage of pixels in the segmentation image that are correctly classified. 
2) Intersection over Union (IoU): The overlap between the character segmentation map in the rendered shot and the character segmentation map in the original shot.
3) Mean Per Joint Position Error (MPJPE): The mean Euclidean distance between the predicted key bone point and the true value.
As shown in Tab.~\ref{tab:constract}, 
(i) Changing backgrounds and moving figures can greatly affect SLAM's accuracy. 
It can be seen from the three metrics that the position of the characters in the picture rendered by SLAM is greatly offset from the original shot, which means that the SLAM method cannot get the correct camera trajectory.
(ii) Since iNeRF only uses RGB pixels as loss, it can not restore the composition of the picture very well.
(iii) Our method achieves the best results in all metrics, 
which shows that our method can accurately restore the composition of the original shot, and the extracted camera trajectory is basically correct.

Fig.~\ref{fig:results_compare} visualizes the rendered image of different methods.
It can be clearly seen that both SLAM and iNeRF rendered images in which character's positions are obviously offset from the original images. 
The images rendered by our method are very consistent, indicating that the camera trajectory we obtained is correct.

\begin{table}[t]
\centering
\resizebox{\columnwidth}{!}{
\begin{tabular}{cc|c|c}
\hline
\multicolumn{2}{c|}{Methods}                                        & SLAM+SMPL & Ours \\ \hline
\multicolumn{1}{c|}{\multirow{2}{*}{2D restoration}} & camera mov. &    4.7$\pm$1.3    &   6.0$\pm$0.5   \\
\multicolumn{1}{c|}{}                                & char. motion &    5.5$\pm$1.1    &   5.8$\pm$0.8   \\ \hline
\multicolumn{1}{c|}{\multirow{2}{*}{3D restoration}} & camera mov. &    4.4$\pm$1.0   &   5.3$\pm$0.6   \\
\multicolumn{1}{c|}{}                                & char. motion &    4.9$\pm$0.9    &    5.0$\pm$1.0  \\ \hline
\end{tabular}
}
\vspace{-8pt}
\caption{
Pair-wise comparison of our method and baseline on content restoration in seven-point Likert scale (lowest-highest:1-7).
}
\label{tab:user_study}
\vspace{-2pt}
\end{table}


\vspace{3pt}
\noindent\textbf{User study.}
To further demonstrate the validity of our method in practice, 
we conduct a user study on 30 shots from different films among 10 volunteers. 
Our study focuses on the accurate recovery of the original video on the screen, and the results of the restored characters and camera movements in world coordinates. 
Volunteers are required to compare the original shot with the generated 2D and 3D results and to determine how well the two matched up with the seven-point Likert scale (lowest-highest:1-7). 
To have a strong baseline, 
we combined DROID-SLAM~\cite{teed2021droid} and SMPL~\cite{ye2023decoupling} estimation to jointly infer camera and character behaviors. 
Volunteers were asked to view the original shot and the results of both methods at the same time, and to rate both results. 
For 2D results, we use the rendered result from camera view like Fig.~\ref{fig:results_jaws}~(c). 
For 3D, we chose a side view that can see the movement of the character and the camera completely like Fig.~\ref{fig:results_smpl}~(d). 
In order to make it easier for volunteers to make judgments, we will provide more than two side views.

Tab.~\ref{tab:user_study} shows that: (i) By employing the NeRF technique to re-optimize the camera trajectory, our method received positive feedback from users who perceiving the extracted camera trajectory as more reasonable compared to SLAM method. 
(ii) Due to the more reasonable camera trajectory to refine SMPL tracks, users have observed enhanced poses in our optimized SMPL, which in turn has created a greater sense of consistency with the original shot characteristics. 
It is important to acknowledge that in movie shots, the character's body is often partially visible, which may lead to an accurate visual representation but lacks accuracy in 3D space. For example, the feet may not be properly positioned on the ground but suspended in the air. So the 3D restoration score is lower compared to the 2D results, considering the additional challenges posed by accurately reconstructing the full 3D context.

\section{Discussion and Conclusion}
\label{sec:conclusion}

In comparison to previous works, our method exhibits improved performance and robustness across a wide range of scenes.
However, our approach still has certain limitations:
i) Our approach relies on a starting camera trajectories obtained from SLAM technology to acquire SMPL tracks. Consequently, when the content of a shot changes too rapidly to extract SMPL tracks, our method is unable to produce the correct results.
ii) Our approach is specifically designed for shots that prominently feature human subjects. 
However, in scenarios where the primary focus of a shot shifts towards showcasing the environment or objects, our method transitions into a simplified version resembling a SLAM approach.

We introduce a reverse filming behavior estimation technique that enables cinematic behavior transfer. 
It utilizes NeRF as a differentiable renderer, effectively optimizes camera trajectories and refines character movements with SMPL models. Additionally, our innovative cinematic transfer pipeline demonstrates its versatility by efficiently transferring various shot types to both 2D video and 3D virtual environments. The integration of a 3D engine workflow not only enhances rendering quality and control but also garners a higher user satisfaction rating, showcasing the potential of our approach in digital media production.
%

\vspace{1pt}
\noindent\textbf{Broader Impact}
\label{sec:broader_impact}
In this paper, we present a method that focuses on transferring cinematic behavior from an existing movie shot to new content, with the aim of enhancing visual storytelling.
With our approach, artists can extract camera trajectory and character movements from shots and use them freely through the engine workflows. This helps them quickly build a prototype for their creation.
It should be noted that this method is made available for academic purposes, and we would like to emphasize that we bear no responsibility for user-generated content. Users are solely accountable for their actions while utilizing this project. The contributors of this project are not legally affiliated with, nor held liable for, the behaviors or actions of users. It is essential to use the method responsibly, adhering strictly to ethical and legal standards.

{
    \small
    \bibliographystyle{ieeenat_fullname}
    \bibliography{main}
}

\end{document}